\documentclass[english]{article}
\usepackage[T1]{fontenc}
\usepackage[utf8]{inputenc}
\usepackage{array}
\usepackage{float}
\usepackage{booktabs}
\usepackage{multirow}
\usepackage{amsmath}
\usepackage{amsthm}
\usepackage{amssymb}
\usepackage{graphicx}

\makeatletter

\providecommand{\tabularnewline}{\\}

\theoremstyle{plain}
\newtheorem{thm}{\protect\theoremname}
\theoremstyle{plain}
\newtheorem{prop}[thm]{\protect\propositionname}
\theoremstyle{plain}
\newtheorem*{prop*}{\protect\propositionname}

\usepackage[nonatbib,preprint]{neurips_2019}

\usepackage{hyperref}       % hyperlinks
\usepackage{url}            % simple URL typesetting
\usepackage{booktabs}       % professional-quality tables
\usepackage{amsfonts}       % blackboard math symbols
\usepackage{nicefrac}       % compact symbols for 1/2, etc.
\usepackage{microtype}      % microtypography

\usepackage{graphicx}

\@ifundefined{showcaptionsetup}{}{%
 \PassOptionsToPackage{caption=false}{subfig}}
\usepackage{subfig}
\makeatother

\usepackage{babel}
\providecommand{\propositionname}{Proposition}
\providecommand{\theoremname}{Theorem}

\begin{document}
\title{Encoder-Powered Generative Adversarial Networks}
\author{Jiseob Kim\\
Seoul National University\\
\texttt{jkim@bi.snu.ac.kr}\\
\And Seungjae Jung\\
Seoul National University\\
\texttt{sjjung@bi.snu.ac.kr}\\
\And Hyundo Lee\\
Seoul National University\\
\texttt{hdlee@bi.snu.ac.kr}\\
\And Byoung-Tak Zhang\\
Seoul National University\\
\texttt{btzhang@bi.snu.ac.kr}\\
}
\maketitle
\begin{abstract}

We present an encoder-powered generative adversarial network (EncGAN)
that is able to learn both the multi-manifold structure and the abstract
features of data. Unlike the conventional decoder-based GANs, EncGAN
uses an encoder to model the manifold structure and invert the encoder
to generate data. This unique scheme enables the proposed model to
exclude discrete features from the smooth structure modeling and learn
multi-manifold data without being hindered by the disconnections.
Also, as EncGAN requires a single latent space to carry the information
for all the manifolds, it builds abstract features shared among the
manifolds in the latent space. For an efficient computation, we formulate
EncGAN using a simple regularizer, and mathematically prove its validity.
We also experimentally demonstrate that EncGAN successfully learns
the multi-manifold structure and the abstract features of MNIST, 3D-chair
and UT-Zap50k datasets. Our analysis shows that the learned abstract
features are disentangled and make a good style-transfer even when
the source data is off the trained distribution.

\end{abstract}

\section{Introduction}

Real-world data involves smooth features as well as discrete features.
The size of a dog would smoothly vary from small to large, but a smooth
transition from a dog to a cat hardly exists as the species is inherently
a discrete feature. Since the discrete feature induces disconnections
in the space, the underlying structure of real-world data is generally
not a single manifold but multiple disconnected manifolds.

Generative adversarial networks (GANs) \cite{goodfellow_generative_2014}
and their successors (e.g., \cite{radford_unsupervised_2015,arjovsky_wasserstein_2017,berthelot_began_2017-1})
present remarkable ability in learning the manifold structure of data.
However, they have difficulties handling the disconnections in multi-manifold
data, since their decoder-based generator primally defines a single
connected manifold \cite{khayatkhoei_disconnected_2018,shao_riemannian_2017,xiao_bourgan_2018}.
Some recent works tackle this issue by using multiple generator networks
\cite{khayatkhoei_disconnected_2018,ghosh_multi-agent_2017,hoang_mgan_2018}
or giving a mixture density on the latent space \cite{xiao_bourgan_2018,gurumurthy_deligan_2017}.
These approaches avoid the difficulties due to the disconnections
as they model each of the manifolds separately.

However, it should be noted that the manifolds generally have shared
structures as they represent common smooth features. For example,
cats and dogs have common smooth features such as the size and the
pose. The respective manifolds of cats and dogs should share their
structures, as the local transforming rule according to the size (stretching
the foreground patch) or the pose (rotating the patches of body parts)
are the same. The separate modeling of manifolds cannot capture this
shared structure, so it cannot learn the common abstract features
of data.

In this work, we propose an encoder-powered GAN (EncGAN) that is able
to learn both the multi-manifold structure and the common abstract
features of data. Unlike the conventional decoder-based GANs, EncGAN
primally uses an encoder for modeling the data. The encoder combines
multiple manifolds into a single latent space by eliminating the disconnected
regions between the manifolds. In the process, the manifolds are aligned
and overlapped, from which the common smooth features are obtained.
Data is generated by inverting the encoder, restoring the eliminated
components to make distinct manifolds. This generating scheme sets
the disconnected regions aside from the modeling of manifolds, thus
resolves the difficulties that GANs have. The advantages of EncGAN
can be summarized as follows:

\begin{itemize}
\item \textbf{Efficient and abstractive modeling:} EncGAN uses a single
encoder to model the multiple manifolds, thus it is efficient and
able to learn the common abstract features.
\item \textbf{Circumvention of modeling the disconnected region:} Discrete
features that induce the disconnections are set aside from the modeling
of manifolds. Thus, EncGAN does not present the difficulties that
GANs have.
\item \textbf{Disentangled features:} Although EncGAN is not explicitly
driven to disentangle the features, it gives a good disentanglement
in the latent-space features due to the shared manifold modeling.
\item \textbf{Easy applicability:} Data generation involves a computationally
challenging inversion of the encoder. However, we propose an inverted,
decoder-like formulation with a regularizer, avoiding the computation
of the inverse. This makes EncGAN easily applicable to any existing
GAN architectures.
\end{itemize}
We start by looking into the difficulties that the conventional GANs
have. Next, we explain our model in detail and investigate the above
advantages with experiments.
\begin{figure}
\centering{}\includegraphics[width=1\textwidth]{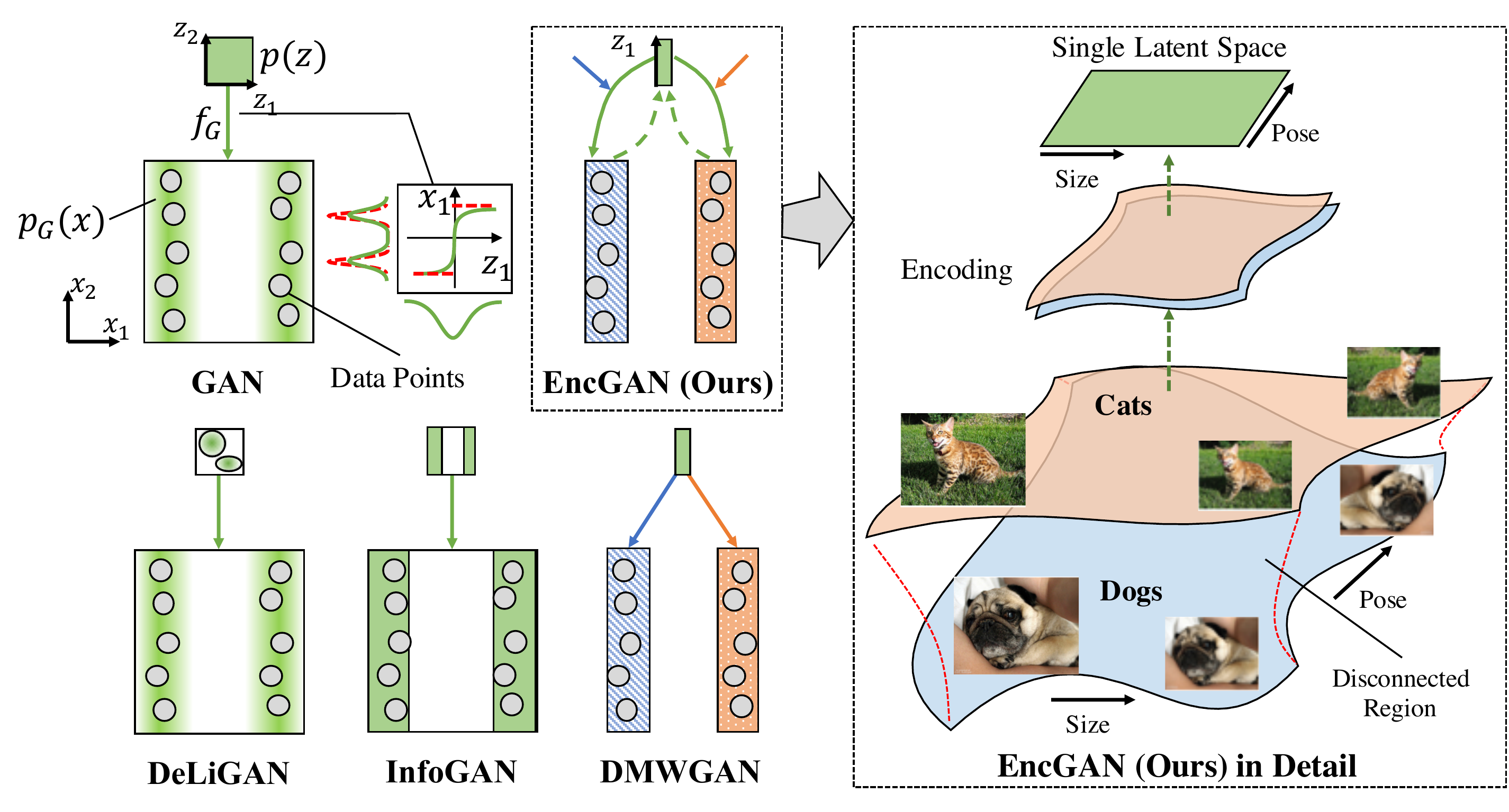}\vspace{-0.5em}
\caption{Various GAN models trained on 2-manifold data are shown schematically.
Green color indicates \emph{shared} elements (functions, distributions)
whereas blue and red indicate \emph{distinct} elements. On the right,
the encoding process of EncGAN is shown in detail. The encoding combines
multiple manifolds into a single latent space by eliminating the disconnected
region. From this, the latent space gets to present the common abstract
features (size and pose). See Sec. \ref{subsec:Separate-Manifold-Modeling}
for more on others.\label{fig:Different-GAN-models}}
\end{figure}

\section{Difficulties in Learning Multiple Manifolds with GAN\label{sec:Difficulties-in-Learning}}

In conventional GANs \cite{goodfellow_generative_2014}, the generator
consists of a latent-space distribution $p(z)$ and a decoding mapping
$f_{G}:Z\to X$. Transforming $p(z)$ using $f_{G}$, an input-space
distribution $p_{G}(x)$ is defined to model the data, and this particular
way of defining $p_{G}(x)$ introduces a manifold structure. The
structure further becomes a single, globally connected manifold, when
$p(z)$ is set to a uniform or a normal distribution as usual. This
is because the supports of aforesaid distributions are globally connected
space, and $f_{G}$ is a smooth and injective\footnote{Although these conditions are not guaranteed, they are approximately
met or observed in practice. See \cite{shao_riemannian_2017}\label{fn:smooth_injective}} map. As smooth, injective maps preserve the connectedness before
and after the mapping, the manifold produced from $f_{G}$ is the
same globally connected space as the latent space \cite{khayatkhoei_disconnected_2018}.

A consequence of this is a difficulty in learning multi-manifold data.
As the generator can only present a single connected manifold, the
best possible option for the generator is to cover all the manifolds
and approximate the disconnected regions with low densities. This
requires the generator to learn a highly non-linear $f_{G}$ because
the density is obtained, using change of variable, as $p_{G}(x)=p(z)/\sqrt{\det\left(J^{\top}J\right)}$
and the Jacobian $J=(\partial f_{G}/\partial z)$ has to be large
to present a low density in $p_{G}(x)$ (see Figure \ref{fig:Different-GAN-models}).
Highly nonlinear functions are hard to learn and often lead to a mode
collapse problem \cite{metz_unrolled_2016,khayatkhoei_disconnected_2018,xiao_bourgan_2018}.
Moreover, even if the generator succeeds in learning all the manifolds,
unrealistic data are sometimes generated as the model presents only
an approximation to the disconnection \cite{khayatkhoei_disconnected_2018}.

\subsection{Separate Manifold Modeling in Extended GANs\label{subsec:Separate-Manifold-Modeling}}

Recently, several extended GAN models are suggested to tackle the
multi-manifold learning. The models can be categorized into two, according
to which component of the generator is extended. The first approach
extends a single decoding mapping $f_{G}$ to multiple decoding mappings
$\{f_{G}^{(i)}\}_{i=1}^{A}$ \cite{khayatkhoei_disconnected_2018,ghosh_multi-agent_2017,hoang_mgan_2018}
(see Figure \ref{fig:Different-GAN-models}, DMWGAN), from which a
disconnected manifold is obtained. In particular, each component distribution
$p_{G}^{(i)}$ obtained from $f_{G}^{(i)}$ models an individual manifold.
The second approach extends the latent-space itself to a disconnected
space, by using Gaussian mixture latent distribution \cite{xiao_bourgan_2018,gurumurthy_deligan_2017}
or by employing discrete latent variables \cite{chen_infogan_2016}
(see Figure \ref{fig:Different-GAN-models}, DeLiGAN and InfoGAN).
Here, each mixture component or discrete variable models an individual
manifold.

Although these extensions work quite well, they could be inefficient
as they consider each manifold separately. Multiple decoding mappings
could require a larger number of parameters, and the disconnected
latent modeling could require much complexity in $f_{G}$, as each
disconnected latent region has to be mapped differently. Also, they
can hardly capture the shared structure that multi-manifold data generally
involves. As we have seen in the cats and dogs example, the manifolds
likely share their structures according to the abstract smooth features.
The separate modeling of manifolds cannot consider this structure,
so they miss learning the abstract features shared among the manifolds.

\section{Encoder-Powered GAN}

Our objective is to model multi-manifold data without having difficulties
due to the disconnections. Also, we aim to model the data efficiently
and capture the abstract features shared among the manifolds. Here,
we explain the proposing encoder-powered GAN in detail and how it
can meet these objectives. We first focus on a single linear layer
then move on to the encoder network inheriting the principles of the
linear layer. Among others, the most important contribution we make
is the inverted formulation using the bias regularizer. We will see
shortly that this enables a tractable model training with mathematical
guarantees.

\begin{figure}
\begin{centering}
\includegraphics[height=0.28\textheight]{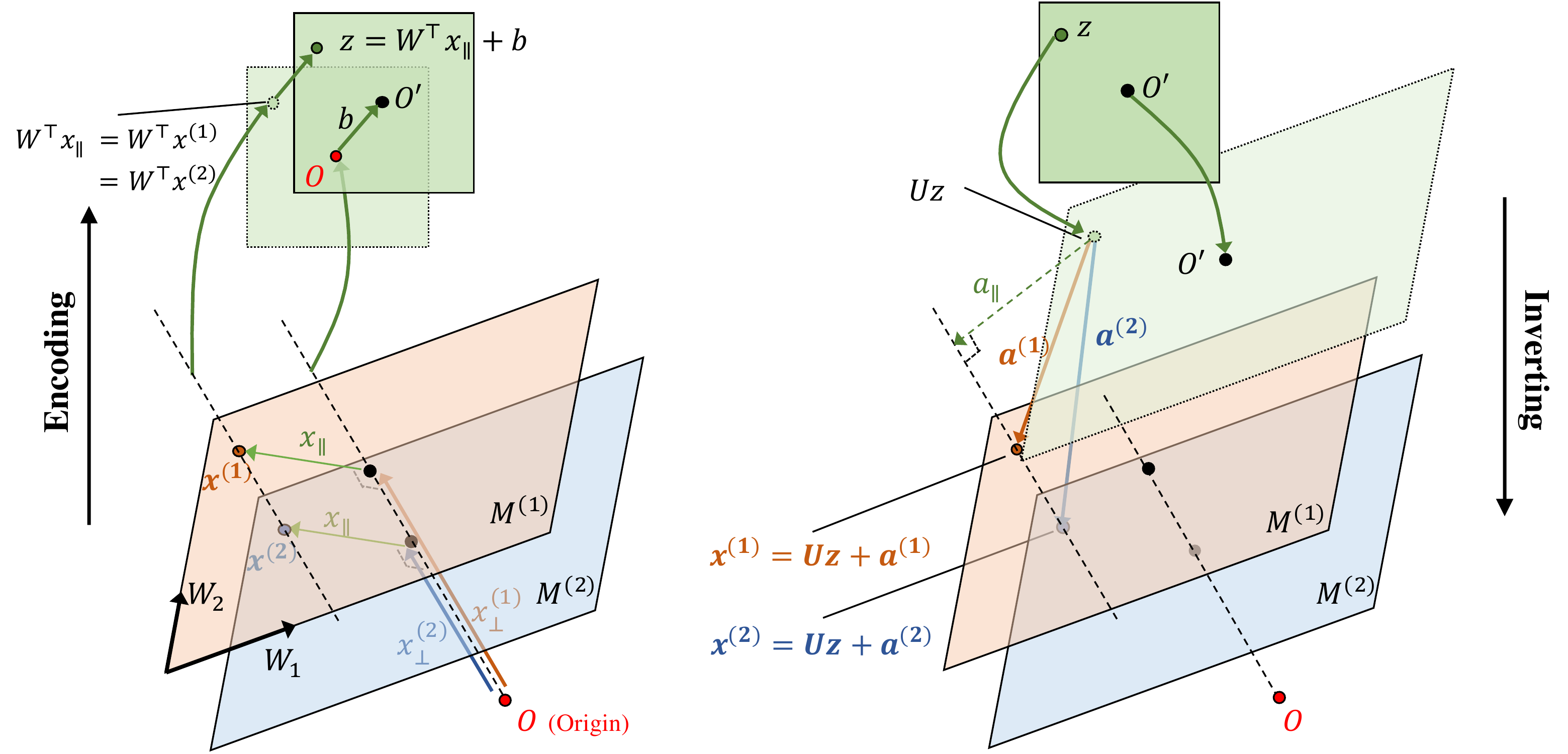}
\par\end{centering}
\caption{Encoding and Inverting in a linear layer. \textbf{Encoding:} Two points
$x^{(1)}$ and $x^{(2)}$ belong to different manifolds but mapped
to the same $z$ as they have the same tangential components $x_{\parallel}$
to the column space of $W$. \textbf{Inverting: }A latent point $z$
is transformed by $U=W(W^{\top}W)^{-1}$, then different decoding
biases $a^{(1)}$ and $a^{(2)}$ are added to generate $x^{(1)}$
and $x^{(2)}$ for different manifolds. The decoding biases are learned
as separated parameters but regularized to have \emph{the} \emph{same
tangential components} to implicitly define the encoder.\label{fig:Encoding-and-Inverting}}
\end{figure}

\subsection{Linear Encoding Layer}

\paragraph{Modeling and Abstracting the Multiple Manifolds}

Consider a linear encoding layer with a weight\footnote{Here, we assume $W$ is full-rank (rank-$d_{z}$) (see Footnote \ref{fn:smooth_injective}
for this assumption).} $W\in\mathbb{R}^{d_{x}\times d_{z}}$ ($d_{z}<d_{x}$) and a bias
$b\in\mathbb{R}^{d_{z}}$. When an input $x\in\mathbb{R}^{d_{x}}$
is passed through the layer, we know from the basic linear algebra
that only the tangential component to the column space of $W$ survives,
and the normal component is lost. We use this property to model the
multiple manifolds and to extract the shared features among them.
In particular, we model the manifolds as spaces parallel to the column
space of $W$, where they differ only in the normal component (see
Figure \ref{fig:Encoding-and-Inverting}, left). This means, a point
lying on the $i$-th manifold can be described as 
\[
x^{(i)}=x_{\parallel}+x_{\perp}^{(i)},
\]
where the tangential component $x_{\parallel}$ is variable and \emph{shared}
among the manifolds and the normal component $x_{\perp}^{(i)}$ is
fixed and \emph{distinct} for each $i$-th manifold. With this modeling,
the latent output 
\begin{equation}
z=W^{\top}x_{\parallel}+b\label{eq:primal_enc}
\end{equation}
is the same for the points lying on different manifolds as long as
they have the same tangential components. This way, the linear layer
captures the smooth features shared among the manifolds in its latent
output. At the same time, it abstracts multiple manifolds into a single
latent space by eliminating the disconnected region along the normal
component.

\paragraph{Data Generation}

Data is generated by inverting the encoding mapping (Eq. \ref{eq:primal_enc}),
from which a random latent sample $z$ is transformed into an input
data sample. Firstly, multiplying the pseudo-inverse of $W$ on the
both sides of Eq. \ref{eq:primal_enc}, we obtain a tangential component
as $x_{\parallel}=W(W^{\top}W)^{-1}(z-b)$. Then, we add a normal
component $x_{\perp}^{(i)}$ to generate the data sample $x^{(i)}$
for each $i$-th manifold:
\begin{equation}
x^{(i)}=W(W^{\top}W)^{-1}(z-b)+x_{\perp}^{(i)}.\label{eq:primal_dec}
\end{equation}
Here, the normal components $\{x_{\perp}^{(i)}\}_{i=1}^{A}$ restore
the disconnections between the manifolds. The normal components can
be either trained or directly computed from the encoder inputs depending
on the model setting. In this work, we only consider the trainable
setting.  

\paragraph{Inverted Formulation}

The primal formulation given above contains the pseudo-inverse of
$W$, which is challenging to compute and unstable for a gradient
descent learning. Our approach here is to construct a dual formulation
based on an inverted, decoder-like parameterization. Although the
model mainly learns decoders under this regime, they are properly
regularized such that the primal single encoder can be recovered anytime.

Let us start by rearranging the Eq. \ref{eq:primal_dec} as
\begin{align*}
x^{(i)} & =Uz+a^{(i)},
\end{align*}
where $U=W(W^{\top}W)^{-1}$ denotes a decoder weight and $a^{(i)}=x_{\perp}^{(i)}-Ub$
denotes a decoder bias for the $i$-th manifold (see Figure \ref{fig:Encoding-and-Inverting},
right). Comparing this rearrangement with a general decoding mapping,
the condition that makes the decoders consistent with the encoder
can be inferred: \textbf{\emph{The weight $U$ and the tangential
component of the bias $a_{\parallel}^{(i)}=-Ub$ are shared among
all the manifolds.}} If we make the decoders keep this condition while
training, we can always recover the encoder by $W=U(U^{\top}U)^{-1}$
and $b=-(U^{\top}U)^{-1}U^{\top}a^{(i)}$. Making $U$ shared is as
trivial as setting the same weight for all the manifolds, but making
$a_{\parallel}^{(i)}$ shared requires a regularization.

\paragraph{Bias Regularizer}

We could make $a_{\parallel}^{(i)}$ shared, by minimizing the sum
of the variance of them: $\text{trace}(\text{cov}(a_{\parallel}^{(i)}))$.
However, computing this term is intractable due to the inversion $(U^{\top}U)^{-1}$
inside of $a_{\parallel}^{(i)}=U(U^{\top}U)^{-1}U^{\top}a^{(i)}$.
\begin{prop}
The following inequality holds
\begin{align*}
\text{trace}\left(\text{cov}(U^{\top}a^{(i)})\right) & \ge\frac{1}{d_{z}}H(\{\lambda_{k}\}_{k=1}^{d_{z}})\text{trace}\left(\text{cov}(a_{\parallel}^{(i)})\right)
\end{align*}
where $\{\lambda_{k}\}_{k=1}^{d_{z}}$ are the eigenvalues of $U^{\top}U$
and $H(\cdot)$ denotes a harmonic mean.
\end{prop}

\begin{proof}
Note that 
\begin{align*}
\text{trace}\left(\text{cov}(a_{\parallel}^{(i)})\right) & =\text{trace}\left(\text{cov}(U(U^{\top}U)^{-1}U^{\top}a^{(i)})\right)\\
 & =\text{trace}\left(\text{cov}(U^{\top}a^{(i)})(U^{\top}U)^{-1}\right)\\
 & \le\text{trace}\left(\text{cov}(U^{\top}a^{(i)})\right)\text{trace}\left((U^{\top}U)^{-1}\right),
\end{align*}
where the second line is obtained from the cyclic property of trace
and the last line is obtained from the Cauchy-Schwarz inequality of
the positive semi-definite matrices (see \ref{apdx:Proposition-1}
for the details).
\end{proof}
As the harmonic mean in the proposition is constant from the perspective
of $a_{\parallel}^{(i)}$, we can minimize the original term by minimizing
the upper bound instead. With an additional $\log$ function to match
the scale due to the dimensionality, we propose the upper bound as
a regularizer for the shared $a_{\parallel}^{(i)}$: 
\[
R_{bias}=\log\left(\text{trace}\left(\text{cov}(U^{\top}a^{(i)})\right)\right)
\]

\subsection{Encoder Network\label{subsec:Encoder-Network}}

A single linear layer is, of course, insufficient as it can only model
the linear manifolds and eliminate the disconnected region only linearly.
So we build a deep encoder network to model the complex nonlinear
manifolds. In each linear layer of the encoder, the disconnected region
is linearly eliminated, so there remains a residual nonlinear region
that is not eliminated. The following nonlinear layer flattens this
region, thus the next linear layer can eliminate it further up to
a smaller number of dimensions. Continuing this procedure layer-by-layer,
the encoder finally yields a single low-dimensional linear space shared
among the manifolds, in which the disconnected region is fully eliminated.

Instead of directly considering the encoder, we again use the inverted
formulation of defining a decoder and regularizing it such that the
encoder can be recovered. Specifically, for each of the linear layers
in the decoder, we set multiple biases and apply the regularizer proposed
above. As the other types of layers (e.g., batch-norm or non-linear
activations) can be inverted in a closed form, we can invert the entire
decoder to recover the original encoder as needed.

\paragraph{Training}

If we denote the decoding weight of the $l$-th linear layer as $U_{l}$
and the decoding biases as $\{a_{l}^{(i)}\}_{i=1}^{A}$ , we can express
the ancestral sampling of data as:
\begin{flalign*}
x & \sim f_{G}^{(i)}\left(z;\{U_{l},a_{l}^{(i)}\}_{l=1}^{L}\right)\quad\text{where}\enskip z\sim p(z),\enskip i\sim\pi_{i}.
\end{flalign*}
Here, $\pi_{i}$ stands for the probability of selecting the $i$-th
bias. This probability could be also learned using the method proposed
in \cite{khayatkhoei_disconnected_2018}, but it is beyond our scope
and we fix it as $1/A$. Now, denoting the real data distribution
as $p_{R}$ and the fake distribution that the above sampling presents
as $p_{G}$, we define our GAN losses as:

\[
\text{\textbf{GAN Loss:} }\begin{cases}
\mathcal{L}_{G}=-E_{x\sim p_{G}}\left[D(x)\right]+\lambda\sum_{l=1}^{L}\log\left(\text{trace}\left(\text{cov}(U_{l}^{\top}a_{l}^{(i)})\right)\right)\\
\mathcal{L}_{D}=E_{x\sim p_{G}}\left[D(x)\right]-E_{x\sim p_{R}}\left[D(x)\right]
\end{cases}
\]
where $\mathcal{L}_{G}$ and $\mathcal{L}_{D}$ are the generator
and the discriminator losses respectively and $\lambda$ is a regularization
weight. We use Wasserstein GAN (WGAN) \cite{arjovsky_wasserstein_2017}
so the discriminator $D(x)$ is limited to a $k$-Lipschitz function.

\paragraph{Encoding}

With the inverted, decoder-like formulation, we have circumvented
the computation of the inversions in data generation; but conversely,
we have difficulties in the encoding, especially due to the convolutional
layers. A recent work considers this difficulty and prove that a distance
minimization approach, $z=\arg\min_{\tilde{z}}\Vert x-f_{G}(\tilde{z})\Vert^{2}$,
can find the correct latent value $z$ for $x$ \cite{ma_invertibility_2018}.
Applying this to our case, we compute the $z$ and the proper decoding
biases by: 
\begin{equation}
z,\{a_{l}\}_{l=1}^{L}=\arg\min_{\tilde{z},\{\tilde{a}_{l}\}_{l=1}^{L}}\left\Vert x-f_{G}\left(\tilde{z};\{\tilde{a}_{l}\}_{l=1}^{L}\right)\right\Vert ^{2}+\mu\sum_{l=1}^{L}\log\left(\left\Vert U_{l}^{\top}a_{l}-U_{l}^{\top}\bar{a}_{l,\parallel}\right\Vert ^{2}\right),\label{eq:encode}
\end{equation}
where the second term is introduced to regularize the shared tangential
component condition, $\mu$ is the regularization weight, and $\bar{a}_{l,\parallel}$
is the mean of the $\{a_{l,\parallel}^{(i)}\}_{i=1}^{A}$.

\section{Experiments}

\begin{table}
\centering{}\caption{FID (smaller is better) and Disentanglement (larger is better) scores
are shown. We compare WGAN \cite{arjovsky_wasserstein_2017}, DMWGAN
\cite{khayatkhoei_disconnected_2018}, $\beta$-VAE \cite{higgins_beta-vae_2016},
InfoGAN \cite{chen_infogan_2016} with our model. The mean and std.
values are computed from 10 (MNIST) and 5 (3D-Chair) replicated experiments.
\label{tab:Scores}}
\scalebox{0.72}{%
\begin{tabular}{cccccccc}
\toprule 
 &  & WGAN & DMWGAN & $\beta$-VAE & InfoGAN & EncGAN (Ours) & EncGAN, $\lambda=0$\tabularnewline
\midrule
\midrule 
\multirow{2}{*}{FID} & MNIST & $10.13\pm3.16$ & $\mathbf{5.41\pm0.34}$ & $58.43\pm0.23$ & $12.17\pm1.30$ & $\mathbf{5.69\pm0.89}$ & $15.74\pm10.00$\tabularnewline
\cmidrule{2-8} \cmidrule{3-8} \cmidrule{4-8} \cmidrule{5-8} \cmidrule{6-8} \cmidrule{7-8} \cmidrule{8-8} 
 & 3D-Chair & $\mathbf{125.32\pm1.16}$ & - & $217.12\pm0.55$ & $187.94\pm9.51$ & \textbf{$\mathbf{125.27\pm4.34}$} & $128.44\pm7.06$\tabularnewline
\midrule 
\multirow{4}{*}{Disent.} & MNIST (slant) & $1.62\pm0.41$ & - & \textbf{$\mathbf{5.04\pm1.19}$} & - & $2.15\pm0.17$ & $1.76\pm0.35$\tabularnewline
\cmidrule{2-8} \cmidrule{3-8} \cmidrule{4-8} \cmidrule{5-8} \cmidrule{6-8} \cmidrule{7-8} \cmidrule{8-8} 
 & MNIST (width) & $1.68\pm0.49$ & - & $\mathbf{5.63\pm0.75}$ & - & $2.93\pm0.60$ & $2.75\pm0.67$\tabularnewline
\cmidrule{2-8} \cmidrule{3-8} \cmidrule{4-8} \cmidrule{5-8} \cmidrule{6-8} \cmidrule{7-8} \cmidrule{8-8} 
 & 3D-Chair (height) & $2.14\pm0.20$ & - & $\mathbf{8.10\pm0.20}$ & - & $3.27\pm1.73$ & $2.76\pm0.31$\tabularnewline
\cmidrule{2-8} \cmidrule{3-8} \cmidrule{4-8} \cmidrule{5-8} \cmidrule{6-8} \cmidrule{7-8} \cmidrule{8-8} 
 & 3D-Chair (bright.) & $3.53\pm0.80$ & - & $3.96\pm0.20$ & - & $\mathbf{4.45\pm0.66}$ & $4.24\pm0.54$\tabularnewline
\bottomrule
\end{tabular}}
\end{table}

\paragraph{Datasets}

We experiment on \emph{MNIST} \cite{lecun_gradient-based_1998}, \emph{3D-Chair}
\cite{aubry_seeing_2014} and \emph{UT-Zap50k} \cite{yu_fine-grained_2014}
image datasets. 3D-Chair contains 1393 distinct chairs rendered for
62 different viewing angles (total 86,366 images); in experiments,
only front-looking 44,576 images are used and rescaled to 64x64 grayscale
images. UT-Zap50k contains images of 4 different types of shoes (total
50,025 images); rescaled to 32x32.

\paragraph{Model Architecture}

We use DCGAN \cite{radford_unsupervised_2015}-like model architectures
for all the datasets (see \ref{apdx:Model-Architecture} for the complete
information). For each of the linear layers in the generator, the
number of biases $A$ are set as 10 (MNIST), 20 (3D-Chair) and 4 (UT-Zap50k).
Although our multi-biased linear layer can be applied to both fully-connected
and transposed-convolution layers, we apply it only to the former.
This is sufficient for our purpose since discrete features rarely
exist for such small sized kernel patches. In the discriminator, we
use a spectral normalization \cite{miyato_spectral_2018} to achieve
the $k$-Lipschitz condition for WGAN. For training and encoding,
Adam \cite{kingma_adam_2014} is used with the defaults except for
the learning rate, 0.0002.

\subsection{Multi-Manifold Learning}

Although the true multi-manifold structures of the datasets are unknown,
we can make a reasonable guess by considering which features are discrete
and which features are smooth. MNIST involves discrete digits and
smoothly varying writing styles, so we can guess it has a distinct
manifold for each digit and the manifolds present the varying writing
styles. Similarly, 3D-Chair would have a distinct manifold for each
category of chairs, and the manifolds represent varying viewing angles,
shapes or brightness; UT-Zap50k would have a distinct manifold for
each type of shoes, and the manifolds represent varying colors or
styles.

\begin{figure}
\includegraphics[width=0.5\textwidth]{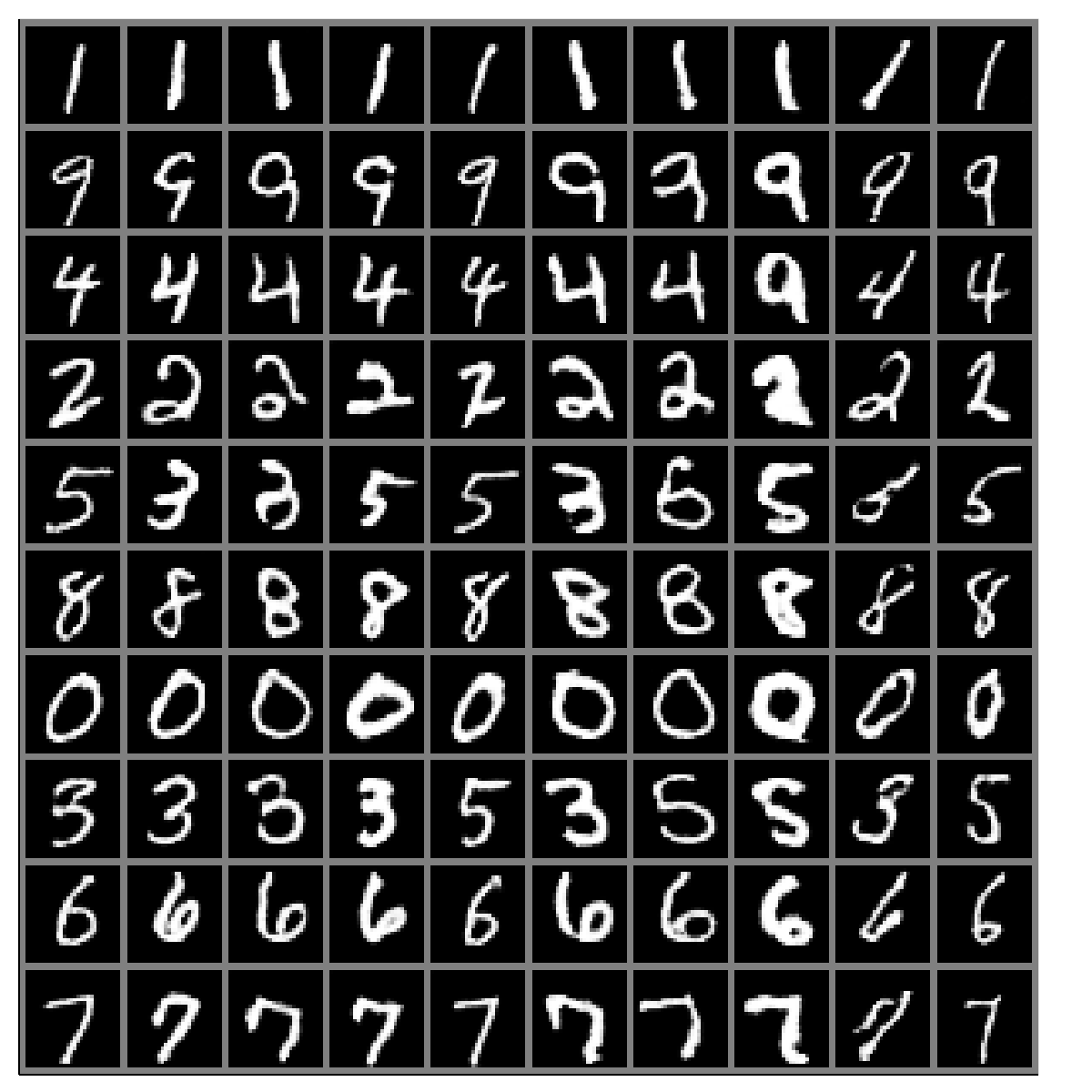}\includegraphics[width=0.5\textwidth]{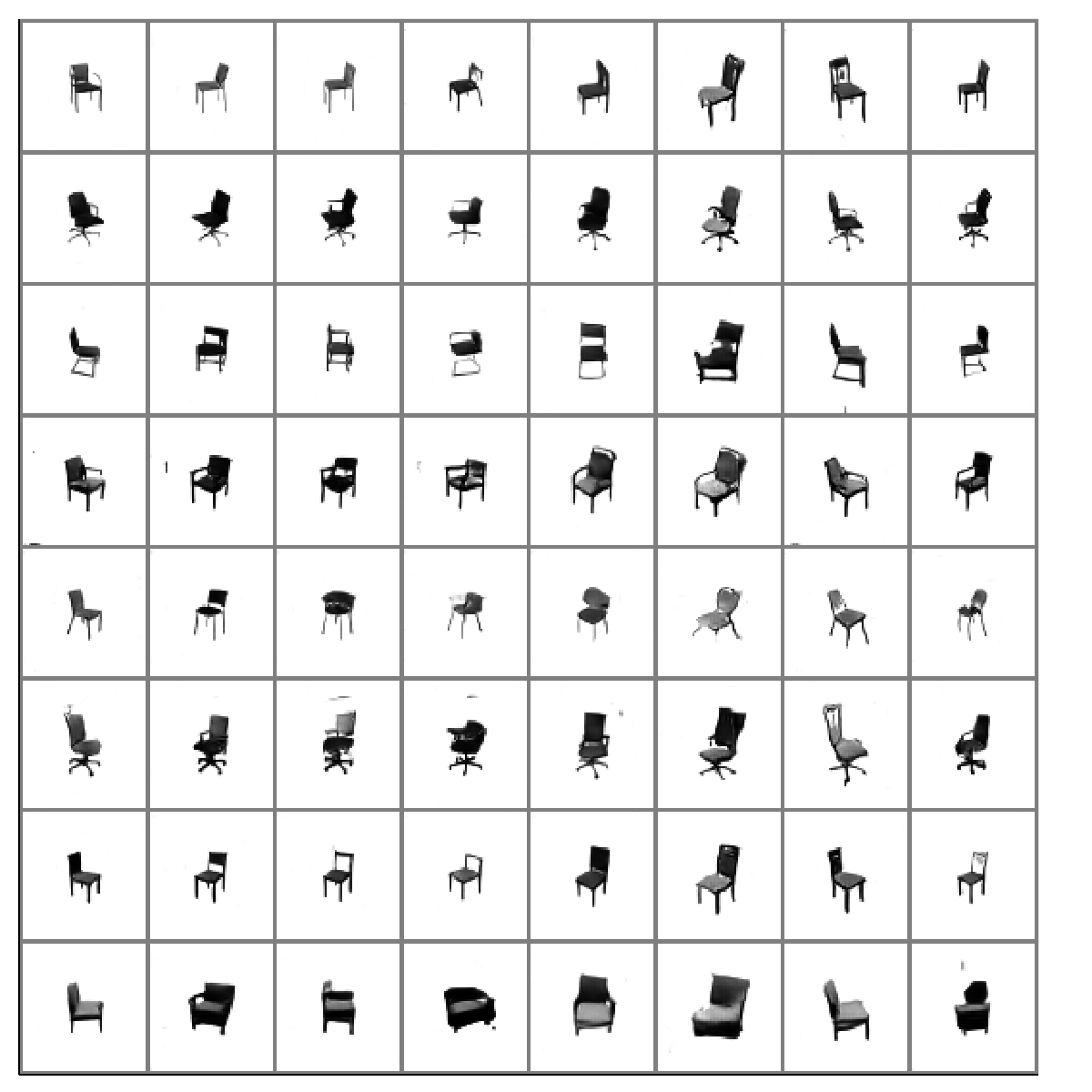}

\includegraphics[width=1\textwidth]{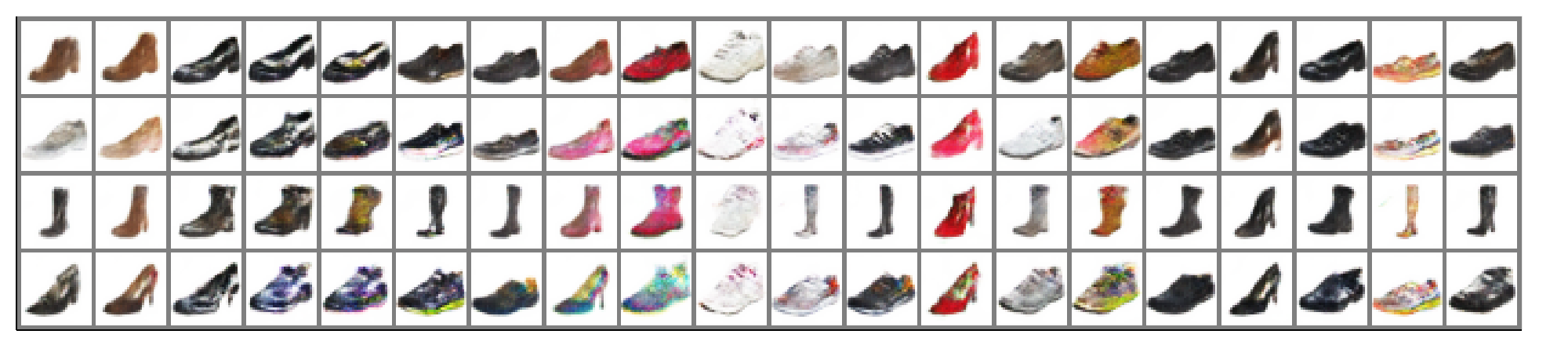}

\caption{Images generated from the trained EncGANs. In all subfigures, rows
indicate distinct manifolds induced by different biases (only 8 out
of 20 are shown for 3D-Chair). The latent codes are randomly sampled
and shared column-wise. \textbf{Row-wise} It can be seen that each
row captures distinct discrete feature, particularly distinct digit
(except 4, 9 and 3, 5 share the manifolds), category of chairs (e.g.,
leg and armrest types) and type of shoes. (e.g., boots and sneakers).
\textbf{Column-wise }It can be seen that smooth features are shared
among the manifolds, particularly stroke weight, slant (MNIST), viewing
angle, height (3D-Chair), and colors (UT-Zap50k). \label{fig:Images-generated-from}}
\end{figure}

Looking at Figure \ref{fig:Images-generated-from} row-wise, we can
see that our model learns distinct manifolds well, in accordance with
our guesses (see in particular the rolling chairs and the boots).
Column-wise, we can see that semantic features (e.g., stroke weight)
are well aligned among the manifolds, which indicates that our model
learns the shared abstract features in the latent space. To quantitatively
examine the sample quality, we compute the FID score \cite{heusel_gans_2017}.
FID score is widely used to quantitatively measure the diversity and
quality of the generated image samples, which also gives a hint of
the presence of a mode collapse. Table \ref{tab:Scores} shows that
our model has better or comparable FID score than others.

\subsection{Disentangled Abstract Features \label{subsec:Disentangled-Abstract-Features}}

\begin{figure}
\begin{centering}
\includegraphics[width=0.2\textwidth]{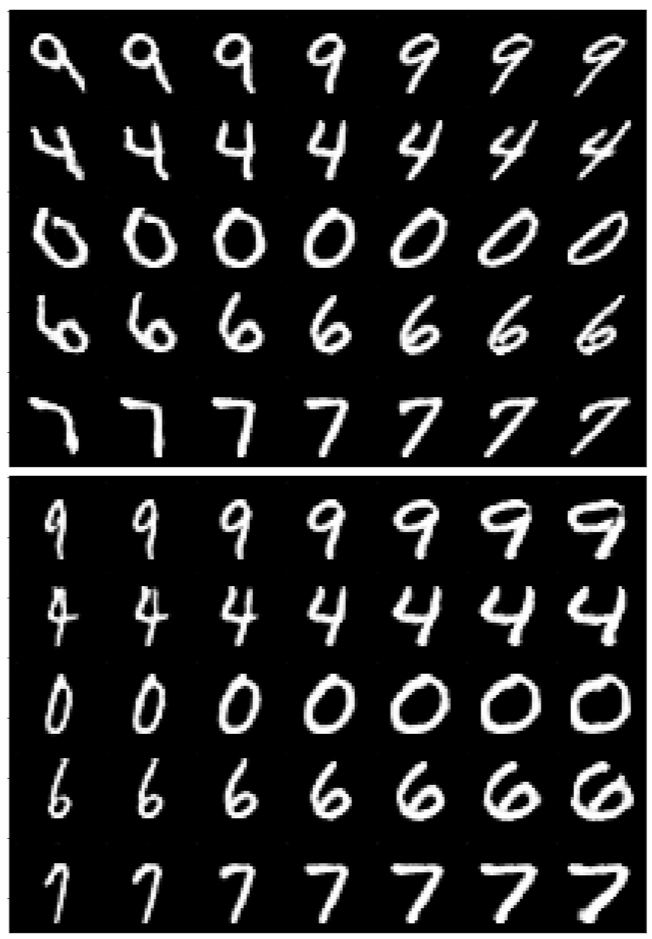}\includegraphics[width=0.4\textwidth]{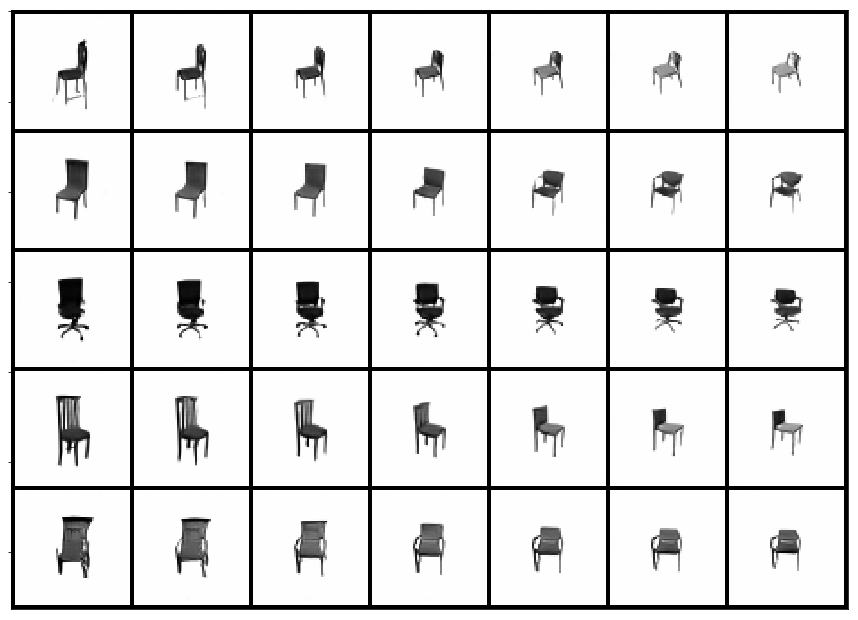}\includegraphics[width=0.4\textwidth]{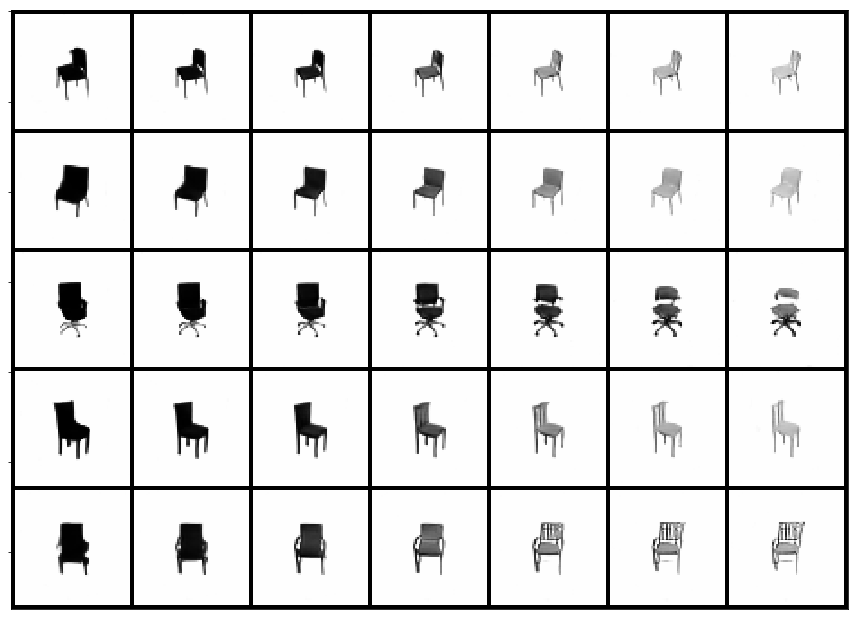}
\par\end{centering}
\caption{Disentangled features. Images are arranged the same as Fig. \ref{fig:Images-generated-from},
except the columns represent linear changes in the latent space along
the first eigenvector in the disentanglement score analysis (see Sec.
\ref{subsec:Disentangled-Abstract-Features}). Slant, width (MNIST),
height and brightness (3D-Chair) components are shown.\label{fig:Disentangled-features}}
\end{figure}

To examine how much our learned latent features are disentangled,
we define a disentanglement score and compare it with other models.
We first take a few images from the dataset and manually change one
of the smooth features that corresponds to a known transformation.
Then we encode these images to the latent codes, analyze the covariance,
and set the ratio of the first eigenvalue to the second as the disentanglement
scores (see \ref{apdx:Disentanglement-Score} for the details). Table
\ref{tab:Scores} shows our model gets better scores than other models,
but sometimes not as good as $\beta$-VAE. But, note that our model
is not guided in information theoretic sense to disentangle the features
as $\beta$-VAE and InfoGAN, yet still shows good disentanglement
as Fig. \ref{fig:Disentangled-features}.

\subsection{Style Transfer}

\begin{figure}
\includegraphics[width=1\textwidth]{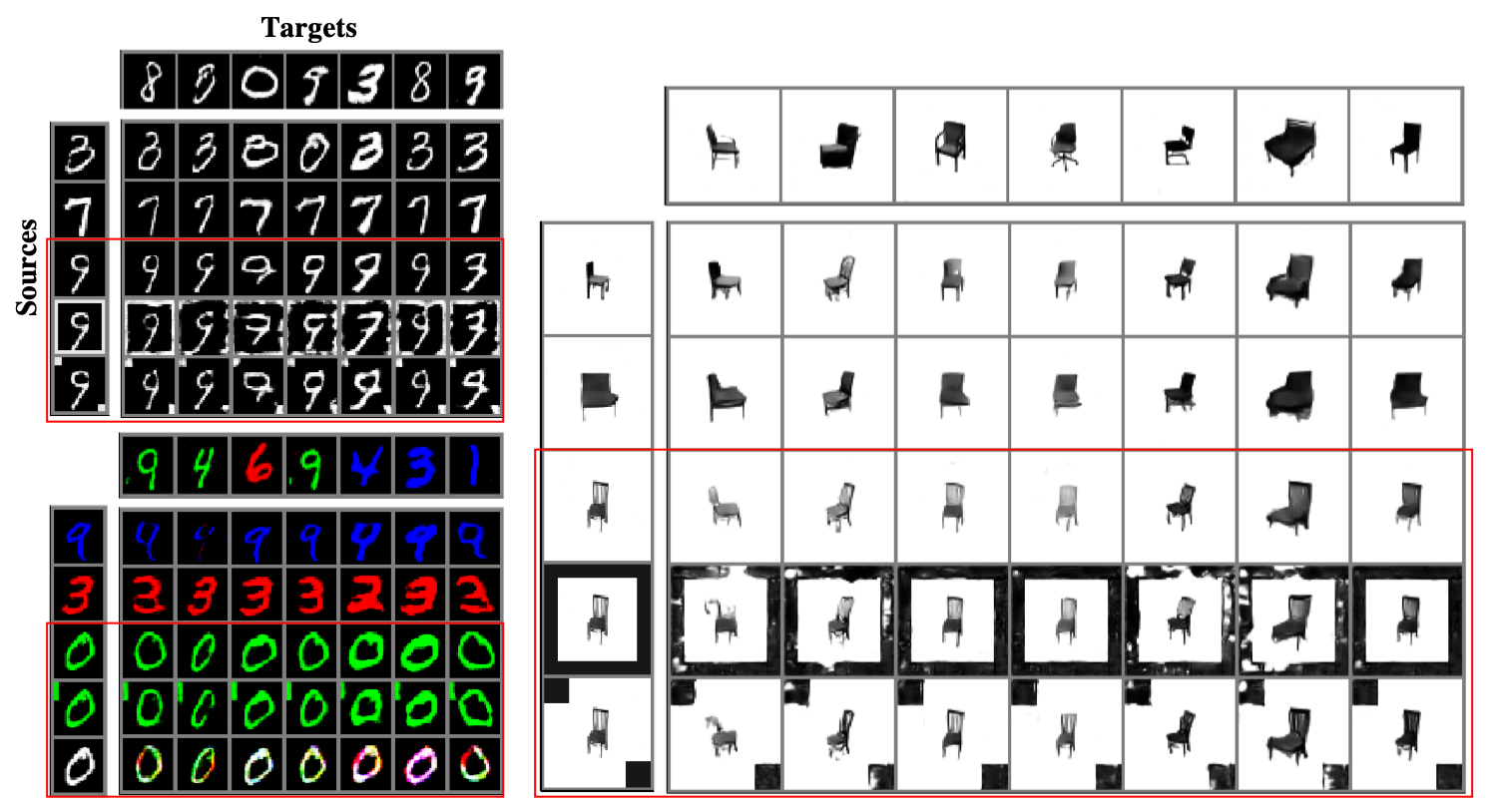}

\caption{Style transfer results. In each source image set, the last two images
are derived from the third image by adding frames, rectangles and
changing colors. The derived source images are clearly off the trained
distribution, but their style-transfer results are quite good and
consistent with the original style-transfer results (highlighted in
red boxes). (For this particular experiment, we trained EncGAN on
RGB-MNIST, a simple extension to MNIST by coloring with either R,
G, or B. We can see that the white image is fairly transferred even
though the model never saw the white color.)\label{fig:Style-transfer-results.}}
\end{figure}

We demonstrate a style-transfer between images using our model, by
matching the latent codes. To make a style-transfer in other models
(e.g., $\beta$-VAE), we usually need to separate the discrete and
smooth features. Then, we transfer only the smooth features such that
the style is transferred, not the identity of the object. In EncGAN,
on the other hand, the latent space contains the smooth features only,
as the discrete features reside in the biases. Thus, we can make style-transferred
images simply by replacing the latent code of the source image with
that of the target images, and regenerate the data (see Fig. \ref{fig:Style-transfer-results.}).

Interestingly, EncGAN is even able to style-transfer the images that
are off the trained distribution (see the red boxes). This is an exclusive
property of EncGAN due to the encoder. If the encoder generalizes
well (which apparently does), it can recognize the abstract features
of an image even if the image has added frame noises or rectangle
noises. Once the features are recognized, the model knows how to transform
this image smoothly just like the original data. So the style-transferred
images can be obtained with the noises still remained, as the noises
are stored in the form of biases (computed from Eq. \ref{eq:encode})
and restored during the regeneration like discrete features.

\section{Conclusion}

In this work, we considered the problem of multi-manifold learning
by proposing encoder-powered GAN (EncGAN). We showed that EncGAN successfully
learns the multi-manifold structure of the data and captures the disentangled
features shared among the manifolds. As EncGAN uses an encoder to
abstractly model the data, it showed a potential to be generalized
to unseen data, as demonstrated in the style-transfer experiment.
If it is trained with larger datasets such as ImageNet in the future,
we expect EncGAN becomes more versatile that it could be able to infer
the manifolds of different datasets (not just the data with noise).
Another line of future work would be transfer-learning the encoder
from the pre-trained classifier models, which would bring a great
boost in GAN training.

\bibliographystyle{plain}
\bibliography{mmgan}

\newpage{}

\appendix
\renewcommand\thesection{Appendix \Alph{section}}

\setcounter{figure}{0} \renewcommand{\thefigure}{\Alph{section}.\arabic{figure}}
\setcounter{table}{0}  \renewcommand{\thetable}{\Alph{section}.\arabic{table}}

\section{Proposition 1 and the Proof\label{apdx:Proposition-1}}
\begin{prop*}
The following inequality holds
\begin{align*}
\text{trace}\left(\text{cov}(U^{\top}a^{(i)})\right) & \ge\frac{1}{d_{z}}H(\{\lambda_{k}\}_{k=1}^{d_{z}})\text{trace}\left(\text{cov}(a_{\parallel}^{(i)})\right)
\end{align*}
where $\{\lambda_{k}\}_{k=1}^{d_{z}}$ are the eigenvalues of $U^{\top}U$
and $H(\cdot)$ denotes a harmonic mean.
\end{prop*}
\begin{proof}
Note that
\begin{align*}
\text{trace}\left(\text{cov}(a_{\parallel}^{(i)})\right) & =\text{trace}\left(\text{cov}(U(U^{\top}U)^{-1}U^{\top}a^{(i)})\right)\\
 & =\text{trace}\left(\frac{1}{A-1}\sum_{i=1}^{A}U(U^{\top}U)^{-1}U^{\top}(a^{(i)}-\bar{a})(a^{(i)}-\bar{a})^{\top}U(U^{\top}U)^{-1}U^{\top}\right)\\
 & =\text{trace}\left(\frac{1}{A-1}\sum_{i=1}^{A}U^{\top}(a^{(i)}-\bar{a})(a^{(i)}-\bar{a})^{\top}U(U^{\top}U)^{-1}\right)\\
 & =\text{trace}\left(\text{cov}(U^{\top}a^{(i)})(U^{\top}U)^{-1}\right)\\
 & \le\text{trace}\left(\text{cov}(U^{\top}a^{(i)})\right)\text{trace}\left((U^{\top}U)^{-1}\right),
\end{align*}
where the second and the fourth lines use the definition of the covariance,
third line is obtained from the cyclic property of trace and the last
line is obtained from the Cauchy-Schwarz inequality of the positive
semi-definite matrices. Thus,
\begin{align*}
\text{trace}\left(\text{cov}(U^{\top}a^{(i)})\right) & \ge\text{trace}\left(\text{cov}(a_{\parallel}^{(i)})\right)/\text{trace}\left((U^{\top}U)^{-1}\right)\\
 & =\frac{1}{D}H(\{\lambda_{d}\}_{d=1}^{D})\text{trace}\left(\text{Var}(a_{\parallel}^{(i)})\right)
\end{align*}
where $\{\lambda_{d}\}_{d=1}^{D}$ are eigenvalues of $U^{\top}U$
and $H(\cdot)$ denotes the harmonic mean.
\end{proof}

\section{Model Architecture and Experimenting Environments\label{apdx:Model-Architecture}}

We used machines with one NVIDIA Titan Xp for the training and the
inference of all the models.

\subsection{MNIST}

We use $A=10$ distinct decoding biases in the model. In the training,
we set the regularization weight $\lambda=0.05$ and use the Adam
optimizer with learning rate 0.0002. In the encoding, we use the Adam
optimizer with learning rate 0.1, and the set the regularization weight
$\mu=0.1$.

\begin{table}[H]
\caption{EncGAN architecture used for MNIST dataset}

\centering{}%
\begin{tabular}{cc}
\toprule 
\textbf{Generator} & \textbf{Discriminator}\tabularnewline
\midrule
\midrule 
Input(8) & Input(1,28,28)\tabularnewline
\midrule 
Full(1024), BN, LReLU(0.2) & Conv(c=64, k=4, s=2, p=1), BN, LReLU(0.2)\tabularnewline
\midrule 
Full(6272), BN, LReLU(0.2) & Conv(c=128, k=4, s=2, p=1), BN, LReLU(0.2)\tabularnewline
\midrule 
ReshapeTo(128,7,7) & ReshapeTo(6272)\tabularnewline
\midrule 
ConvTrs(c=64, k=4, s=2, p=1), BN, LReLU(0.2) & Full(1024), BN, LReLU(0.2)\tabularnewline
\midrule 
ConvTrs(c=32, k=4, s=2, p=1), BN, LReLU(0.2) & Full(1)\tabularnewline
\midrule 
ConvTrs(c=1, k=3, s=1, p=1), Tanh & \tabularnewline
\bottomrule
\end{tabular}
\end{table}

\subsubsection{Notes on the Other Compared Models}

Overall, we match the architecture of other models with our model
for fair comparison. Some differences to note are:
\begin{itemize}
\item \textbf{DMWGAN}: We used 10 generators. Each generator has the same
architecture as ours except the number of features or the channels
are divided by 4, to match the number of trainable parameters. Note
that 4 is the suggested number from the original paper.
\item \textbf{$\beta$-VAE}: We used Bernoulli likelihood.
\item \textbf{InfoGAN}: Latent dimensions consist of 1 discrete variable
(10 categories), 2 continuous variable and 8 noise variable.
\end{itemize}

\subsection{3D-Chair}

We use $A=20$ distinct decoding biases in the model. In the training,
we set the regularization weight $\lambda=0.05$ and use the Adam
optimizer with learning rate 0.0002. In the encoding, we use the Adam
optimizer with learning rate 0.1, and the set the regularization weight
$\mu=0.1$.

\begin{table}[H]
\caption{EncGAN architecture used for 3D-Chair dataset.}

\centering{}%
\begin{tabular}{cc}
\toprule 
\textbf{Generator} & \textbf{Discriminator}\tabularnewline
\midrule
\midrule 
Input(10) & Input(1,64,64)\tabularnewline
\midrule 
Full(256), BN, LReLU(0.2) & Conv(c=64, k=4, s=2, p=1), BN, LReLU(0.2)\tabularnewline
\midrule 
Full(8192), BN, LReLU(0.2) & Conv(c=128, k=4, s=2, p=1), BN, LReLU(0.2)\tabularnewline
\midrule 
ReshapeTo(128,8,8) & Conv(c=128, k=4, s=2, p=1), BN, LReLU(0.2)\tabularnewline
\midrule 
ConvTrs(c=64, k=4, s=2, p=1), BN, LReLU(0.2) & ReshapeTo(8192)\tabularnewline
\midrule 
ConvTrs(c=32, k=4, s=2, p=1), BN, LReLU(0.2) & Full(1024), BN, LReLU(0.2)\tabularnewline
\midrule 
ConvTrs(c=16, k=4, s=2, p=1), BN, LReLU(0.2) & Full(1)\tabularnewline
\midrule 
ConvTrs(c=1, k=3, s=1, p=1), Tanh & \tabularnewline
\bottomrule
\end{tabular}
\end{table}

\subsubsection{Notes on the Other Compared Models}
\begin{itemize}
\item \textbf{$\beta$-VAE}: We used Bernoulli likelihood.
\item \textbf{InfoGAN}: Latent dimensions consist of 3 discrete variable
(20 categories), 1 continuous variable and 10 noise variable.
\end{itemize}

\subsection{UT-Zap50k}

We use $A=4$ distinct decoding biases in the model. For the regularization
weight in the training, we start with $\lambda=5e-6$ then raise to
$\lambda=5e-4$ after 300 epochs.

\begin{table}[H]
\caption{EncGAN architecture used for UT-Zap50k dataset.}

\centering{}%
\begin{tabular}{cc}
\toprule 
\textbf{Generator} & \textbf{Discriminator}\tabularnewline
\midrule
\midrule 
Input(8) & Input(3,32,32)\tabularnewline
\midrule 
Full(512), BN, LReLU(0.2) & Conv(c=128, k=4, s=2, p=1), BN, LReLU(0.2)\tabularnewline
\midrule 
Full(1024), BN, LReLU(0.2) & Conv(c=256, k=4, s=2, p=1), BN, LReLU(0.2)\tabularnewline
\midrule 
Full(8192), BN, LReLU(0.2) & Conv(c=512, k=4, s=2, p=1), BN, LReLU(0.2)\tabularnewline
\midrule 
ReshapeTo(512,4,4) & ReshapeTo(8192)\tabularnewline
\midrule 
ConvTrs(c=256, k=4, s=2, p=1), BN, LReLU(0.2) & Full(1024), BN, LReLU(0.2)\tabularnewline
\midrule 
ConvTrs(c=128, k=4, s=2, p=1), BN, LReLU(0.2) & Full(512), BN, LReLU(0.2)\tabularnewline
\midrule 
ConvTrs(c=64, k=4, s=2, p=1), BN, LReLU(0.2) & Full(1)\tabularnewline
\midrule 
ConvTrs(c=3, k=3, s=1, p=1), Tanh & \tabularnewline
\bottomrule
\end{tabular}
\end{table}

\section{Disentanglement Score\label{apdx:Disentanglement-Score}}

To compute the disentanglement score, we first take 500 images from
the dataset and manually change one of the smooth features that corresponds
to a known transformation. For example, we change the slant of the
MNIST digits by taking a sheer transform. With 11 different degrees
of the transformation, we obtain 5500 transformed images in total.
We encode these images to obtain the corresponding latent codes and
subtract the mean for each group of the images (originates from the
same image) to align all the latent codes. Then, we conduct Principal
Component Analysis (PCA) to obtain the principal direction and the
spectrum of variations of the latents codes. If the latent features
are well disentangled, the dimensionality of the variation should
be close to one. To quantify how much it is close to one, we compute
the ratio of the first eigenvalue to the second eigenvalue of the
PCA covariance, and set it as the disentanglement score.

\section{Effect of the Bias Regularizer}

\begin{figure}
\begin{centering}
\includegraphics[width=0.5\textwidth]{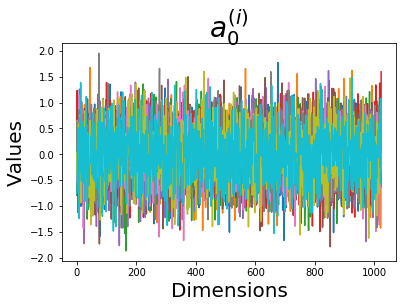}\includegraphics[width=0.5\textwidth]{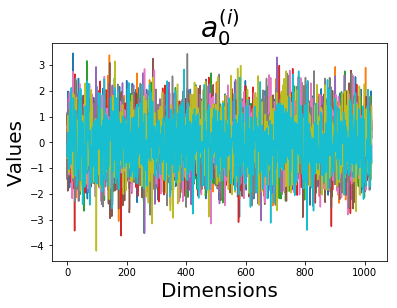}
\par\end{centering}
\begin{centering}
\includegraphics[width=0.5\textwidth]{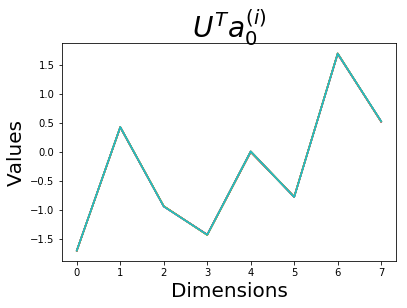}\includegraphics[width=0.5\textwidth]{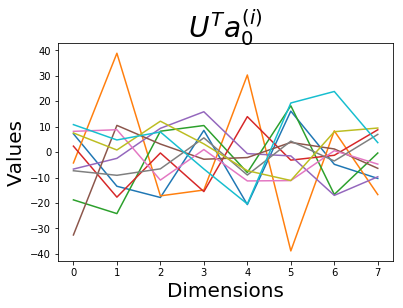}
\par\end{centering}
\begin{centering}
\includegraphics[width=0.5\textwidth]{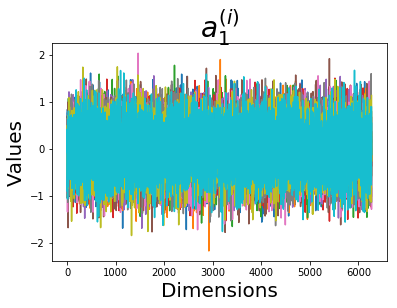}\includegraphics[width=0.5\textwidth]{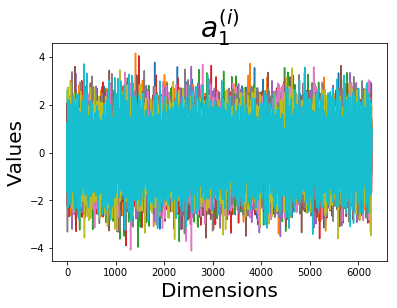}
\par\end{centering}
\begin{centering}
\includegraphics[width=0.5\textwidth]{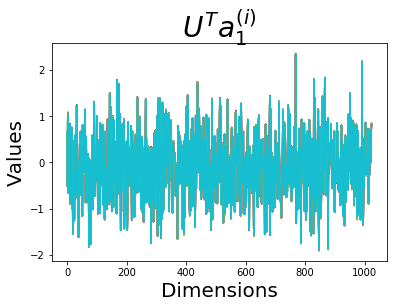}\includegraphics[width=0.5\textwidth]{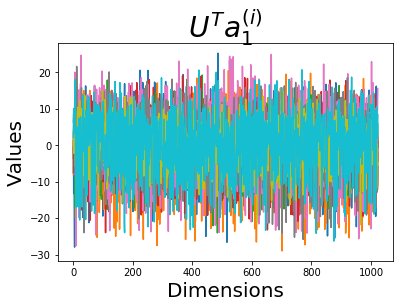}
\par\end{centering}
\centering{}\caption{Biases $a_{l}^{(i)}$ and their (pseudo-)tangential component $U_{l}^{\top}a_{l}^{(i)}$
of the EncGAN models, trained on \textbf{\emph{MNIST}}. Individual
curve indicates each $i$-th bias.\textbf{ Left }Parameters of EncGAN
\textbf{Right} Parameters of EncGAN without thes regularizer ($\lambda=0$).
It can be seen that the regularizer makes the tangential components
of the biases well aligned. \label{fig:Biases-MNIST}}
\end{figure}

\begin{figure}
\begin{centering}
\includegraphics[width=0.5\textwidth]{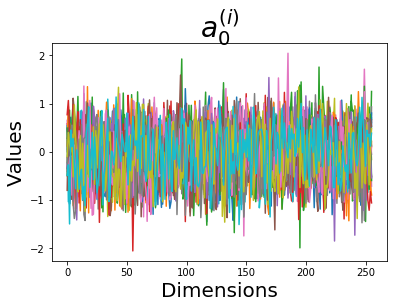}\includegraphics[width=0.5\textwidth]{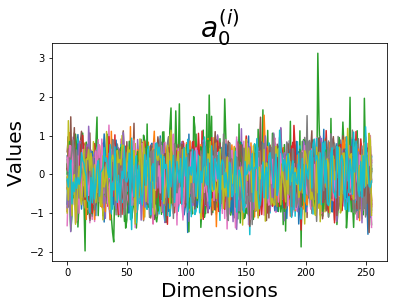}
\par\end{centering}
\begin{centering}
\includegraphics[width=0.5\textwidth]{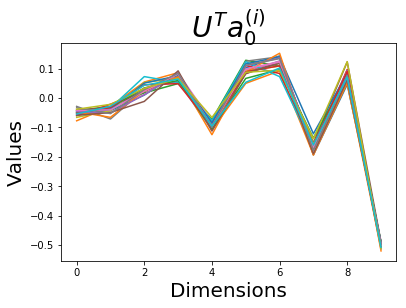}\includegraphics[width=0.5\textwidth]{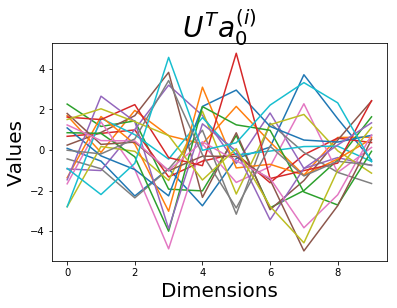}
\par\end{centering}
\begin{centering}
\includegraphics[width=0.5\textwidth]{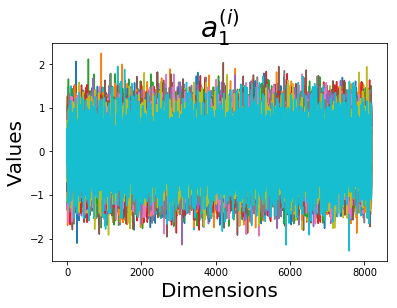}\includegraphics[width=0.5\textwidth]{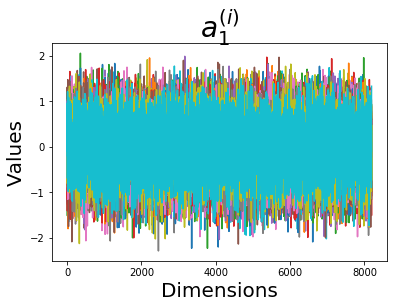}
\par\end{centering}
\begin{centering}
\includegraphics[width=0.5\textwidth]{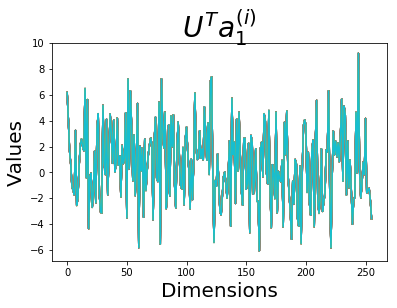}\includegraphics[width=0.5\textwidth]{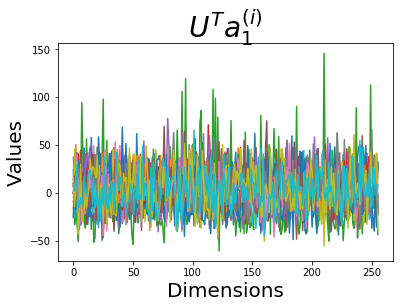}
\par\end{centering}
\centering{}\caption{Biases $a_{l}^{(i)}$ and their (pseudo-)tangential component $U_{l}^{\top}a_{l}^{(i)}$
of the EncGAN models, trained on \textbf{\emph{3D-Chair}}. Individual
curve indicates each $i$-th bias.\textbf{ Left }Parameters of EncGAN
\textbf{Right} Parameters of EncGAN without the regularizer ($\lambda=0$).
It can be seen that the regularizer makes the tangential components
of the biases well aligned.\label{fig:Biases-Chair}}
\end{figure}

To examine the effectiveness of our bias regularizer, we visualize
the raw values of biases $\{a_{l}^{(i)}\}_{i,l}$ and their (pseudo-)tangential
component $\{U_{l}^{\top}a_{l}^{(i)}\}_{i,l}$ (see Fig. \ref{fig:Biases-MNIST},
\ref{fig:Biases-Chair}). In all figures, we see that the biases are
diverse, but their tangential components are well aligned due to the
bias regularizer (left). On the contrary, without the regularizer,
the tangential components are not aligned (right).

\section{Samples Generated from Various Models}

\begin{figure}
\subfloat[EncGAN]{\begin{centering}
\includegraphics[width=0.45\textwidth]{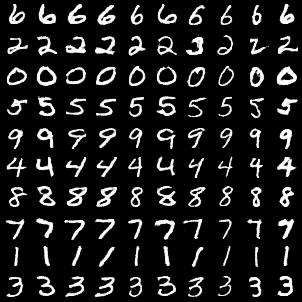}
\par\end{centering}
}\subfloat[EncGAN ($\lambda=0$)]{\begin{centering}
\includegraphics[width=0.45\textwidth]{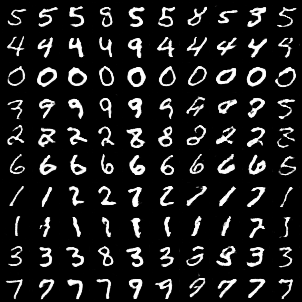}
\par\end{centering}
}

\subfloat[DMWGAN]{\begin{centering}
\includegraphics[width=0.45\textwidth]{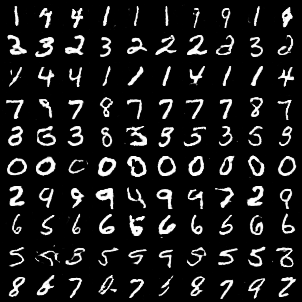}
\par\end{centering}
}\subfloat[WGAN]{\begin{centering}
\includegraphics[width=0.45\textwidth]{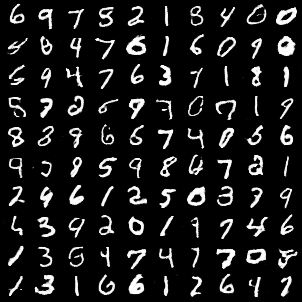}
\par\end{centering}
}

\subfloat[$\beta$-VAE]{\begin{centering}
\includegraphics[width=0.45\textwidth]{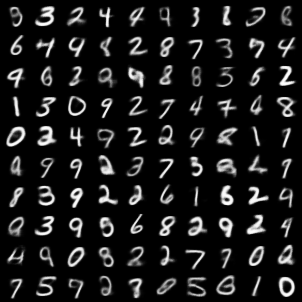}
\par\end{centering}
}\subfloat[InfoGAN]{\begin{centering}
\includegraphics[width=0.45\textwidth]{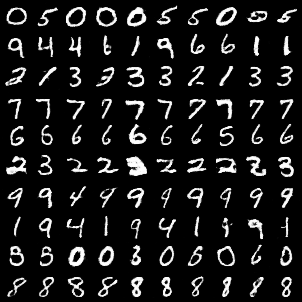}
\par\end{centering}
}

\caption{MNIST image samples generated from the trained models}
\end{figure}

\begin{figure}
\begin{centering}
\subfloat[WGAN]{\begin{centering}
\includegraphics[height=0.65\paperheight]{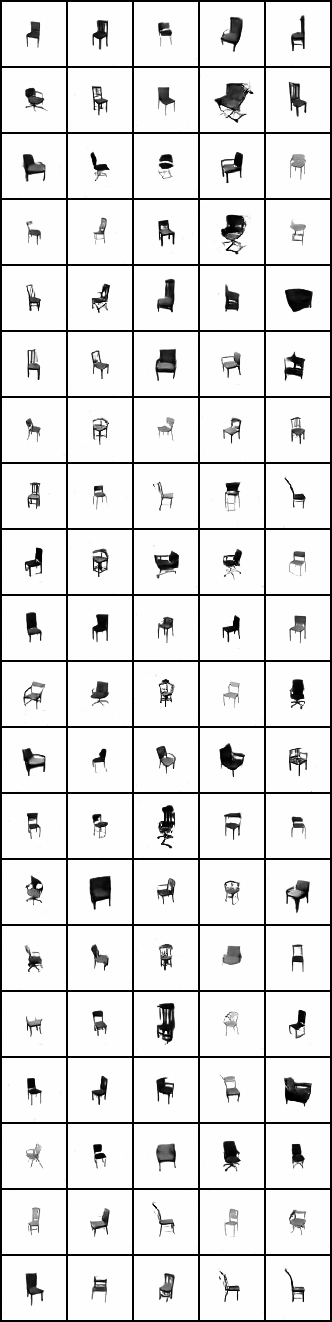}
\par\end{centering}
}\subfloat[$\beta$-VAE]{\begin{centering}
\includegraphics[height=0.65\paperheight]{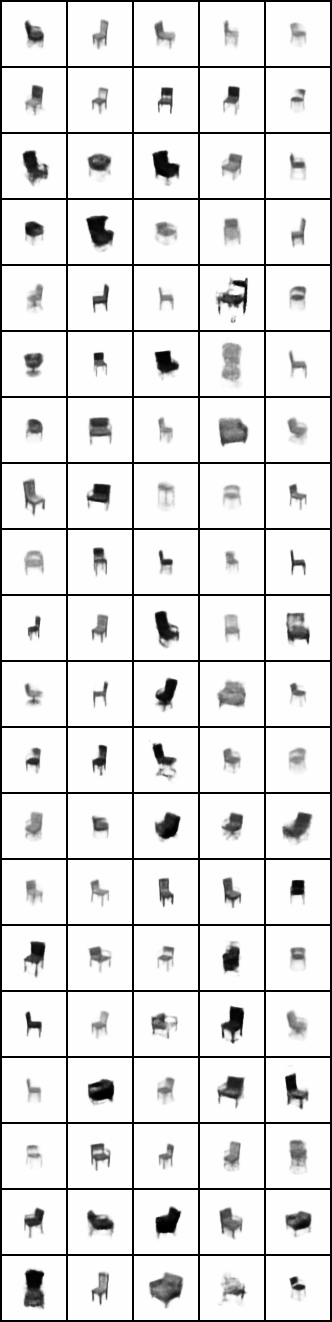}
\par\end{centering}
}\subfloat[InfoGAN]{\begin{centering}
\includegraphics[height=0.65\paperheight]{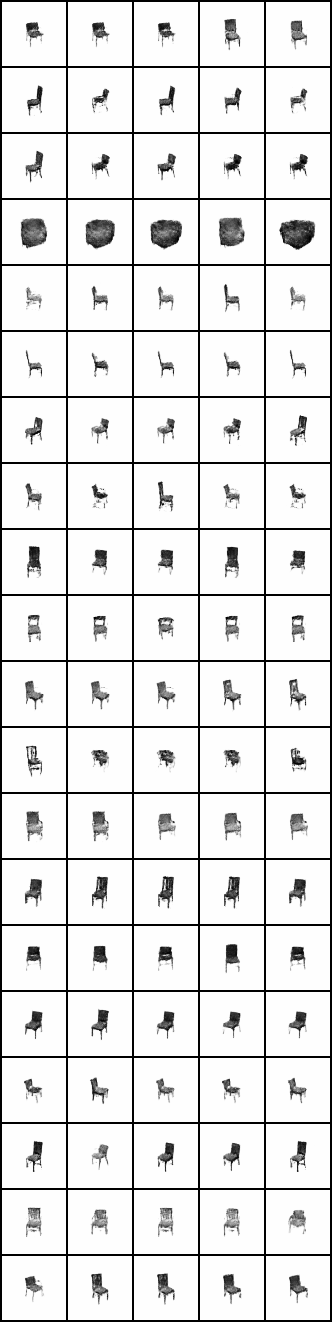}
\par\end{centering}
}
\par\end{centering}
\caption{3D-Chair image samples generated from the trained models}
\end{figure}

\begin{figure}
\begin{centering}
\includegraphics[height=0.7\paperheight]{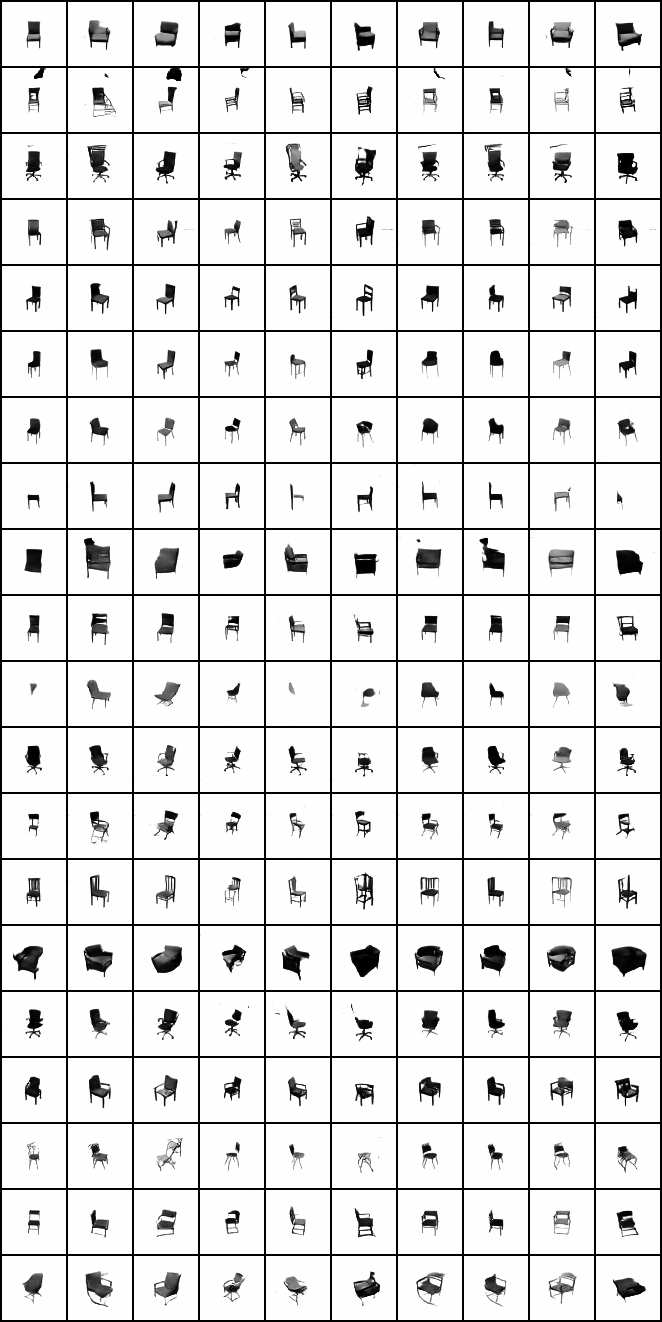}
\par\end{centering}
\caption{3D-Chair image samples generated from the trained EncGAN}
\end{figure}

\begin{figure}
\begin{centering}
\includegraphics[height=0.7\paperheight]{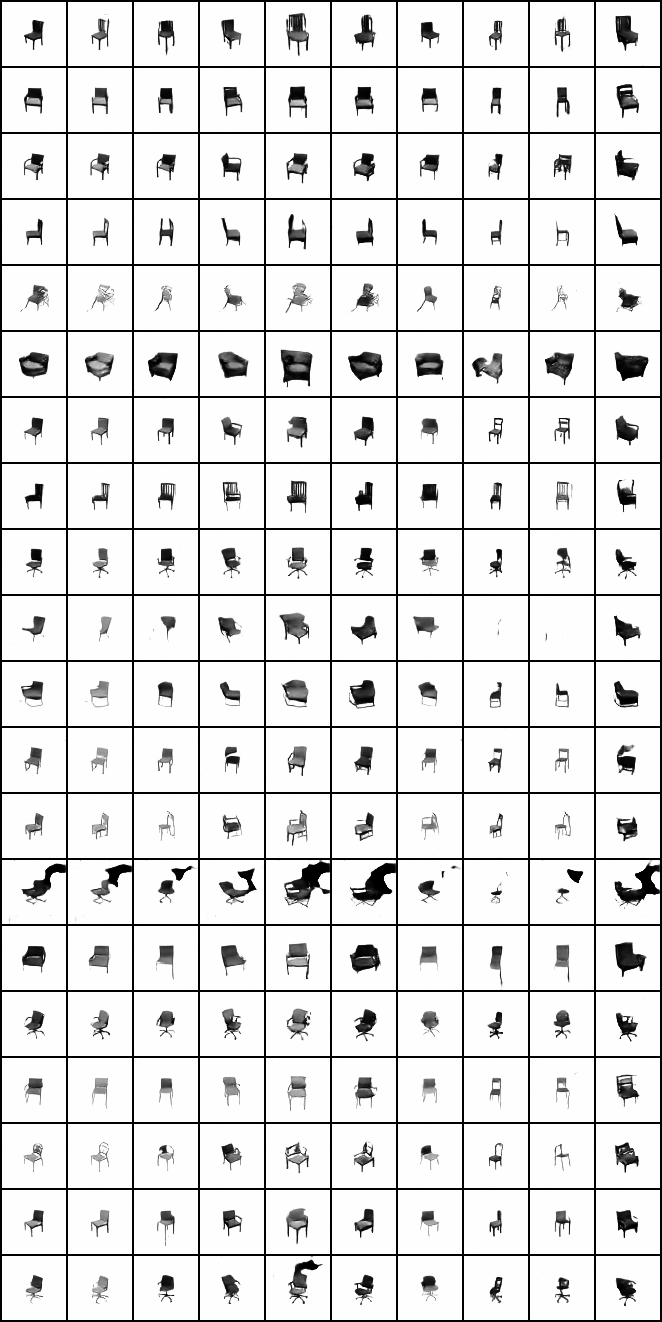}
\par\end{centering}
\caption{3D-Chair image samples generated from EncGAN ($\lambda=0$)}
\end{figure}

\end{document}